\documentclass[10pt, final, letter, twocolumns]{IEEEtran}
\usepackage{wrapfig}
\usepackage{amssymb}
\usepackage{latexsym}
\usepackage{amsmath}
\usepackage{alltt}
\usepackage{graphicx}
\usepackage{amsfonts}
\usepackage{color,colortbl}
\usepackage{float}
\usepackage{multicol}
\usepackage{multirow}
\usepackage{hhline}
\usepackage{graphicx}
\usepackage{booktabs}
\usepackage{rotating}
\usepackage{algorithm}
\usepackage{algorithmicx}
\usepackage{algpseudocode}
\usepackage{tabto}
\usepackage{lipsum}
\usepackage{siunitx}
\usepackage[flushleft]{threeparttable}
\usepackage{xcolor}
\usepackage{subfigure}
\usepackage[bookmarks=false, hidelinks]{hyperref}
\usepackage{pdfpages}
\usepackage{cite}
\usepackage[T1]{fontenc}

\newcommand{\exclude}[1]{}

\newcommand{\bU}{\textbf{U}}

\newcommand{\bV}{\textbf{V}}

\newcommand{\bR}{\textbf{R}}
\newcommand{\bP}{\textbf{P}}
\newcommand{\bQ}{\textbf{Q}}
\newcommand{\bDelta}{\boldsymbol\Sigma}

\definecolor{Gray}{gray}{0.9}
\definecolor{LightCyan}{rgb}{0.88,1,1}

\algdef{SE}[DOWHILE]{Do}{doWhile}{\algorithmicdo}[1]{\algorithmicwhile\ #1}

\usepackage[normalem]{ulem}

\begin{document}

    \title{Efficient Inference of CNNs via Channel Pruning}
    \author{}
        \author{
        \IEEEauthorblockN{Boyu Zhang, Azadeh Davoodi, and Yu Hen Hu}\\
        \IEEEauthorblockA{University of Wisconsin--Madison\\
                          1415 Engineering Drive, Madison, WI, USA 53706\\
                          \{bzhang93, adavoodi, yhhu\}@wisc.edu}
        }
    \maketitle
    \hyphenpenalty=3000
    
    \begin{abstract}
    The deployment of Convolutional Neural Networks (CNNs) on resource constrained platforms such as mobile devices and embedded systems has been greatly hindered by their high implementation cost, and thus motivated a lot research interest in compressing and accelerating trained CNN models. Among various techniques proposed in literature, structured pruning, especially channel pruning, has gain a lot focus due to 1) its superior performance in memory, computation, and energy reduction; and 2) it is friendly to existing hardware and software libraries. In this paper, we investigate the intermediate results of convolutional layers and present a novel pivoted QR factorization based channel pruning technique that can prune any specified number of input channels of any layer. We also explore more pruning opportunities in ResNet-like architectures by applying two tweaks to our technique. Experiment results on VGG-16 and ResNet-50 models with ImageNet ILSVRC 2012 dataset are very impressive with 4.29X and 2.84X computation reduction while only sacrificing about 1.40\% top-5 accuracy. Compared to many prior works, the pruned models produced by our technique require up to 47.7\% less computation while still achieve higher accuracies. 
\end{abstract}
    \section{Introduction}
Deep Convolutional Neural Networks (CNNs) have achieved great success in computer vision related tasks such as image classification\cite{vgg}, object detection\cite{objectdetection}, semantic segmentation\cite{segmentation}, image captioning\cite{captioning}, and so on. Besides the availability of large-scale datasets and advances of modern GPUs, increasingly more complicated model architecture is one of the most important reasons that consistently pushing the performance of CNNs to higher level. However, complicated models often contain millions of parameters and require billions of operations per inference, which hinders its deployment on resource constrained platforms such as mobile devices and embedded systems.

Specifically, the difficulties of deploying CNNs on resource constrained devices are coming from four aspects: 
\textbf{1) Static memory:} Modern CNNs for practical applications often contain millions of parameters, which need to be loaded into memory during inference. Although the memory size of embedded devices is increasing in recent years, large-size CNNs still impose a heavy burden on memory, especially when considering other applications running simultaneously;
\textbf{2) Run-time memory:} Besides model parameters, intermediate activation values also need to be stored in memory during inference. This type of memory requirement highly depends on batch size and whether gradient back propagation will be performed. Although it can be alleviated by reducing batch size at the cost of sacrificing throughput, runtime memory is not negligible even if batch size equals to 1; 
\textbf{3) Computation capability:} Billions of computations are often required per inference, and they need to be accomplished in short time to deliver satisfactory user experience. This is a very challenging task for embedded processors due to limited computation capability and other related factors, such as cache effects, memory accessing, operating system;
\textbf{4) Energy consumption:} Conducting billions of computations and accessing (load and store) millions model parameters/intermediate activations are very energy-consuming. Thus, for devices that are powered by battery, battery life may become an issue. 

To address these difficulties, many model compression techniques (low-rank approximation\cite{lowrank}, quantization\cite{quantization}, weight pruning\cite{deepcompression}, and etc.) and more efficient network architectures (MobileNet\cite{mobilenets}, ShuffleNet\cite{shufflenet}, and etc.) have been proposed. Pruning techniques have been studied since 90s\cite{braindamage, brainsurgeon} and can be further divided into unstructured pruning and structured pruning. As a representative unstructured pruning technique, \cite{deepcompression} pruned weights and neurons whose values are below a certain magnitude and demonstrated very good theoretical compression ratio and speedup results. However, like many other unstructured pruning techniques, the effects of pruning are reside in the sparsity of weight matrices rather than the network architecture. Thus, the benefits of the resulting pruned model cannot be easily realized without the support of specialized hardware and software. 

In contrast, structured pruning aims to prune models in a structured manner, such as pruning entire layers or pruning channels inside convolutional layers. Thus, the resulting pruned models become more efficient in terms of network architecture, which makes the theoretical benefits brought by structured pruning techniques easier to be realized by existing hardware and deep learning libraries.

\textbf{In this paper}, we propose a structured pruning technique that identifies and prunes redundant input channels of convolutional layers. After a channel is pruned, its corresponding kernel weights of both current layer and previous layer will be pruned as well, thus achieving significant reduction in terms of static and run-time memory, computation, and energy consumption. The proposed procedure investigates the intermediate results of convolutional layers and formulate the task of identifying redundant channels as a subset selection problem, in which a subset of input channels of specified size is selected to maximally preserve the original outputs of the target layer and the other channels are identified as redundant. This NP-hard problem is approximately solved by adopting a well-known efficient algorithm utilizing QR factorization with column pivoting. We also propose two tweaks for the procedure to explore more pruning opportunity in ResNet-like models. Moreover, the proposed technique is orthogonal to others such as quantization and low-rank expansion, which can be combined to achieve further reduction.

To validate the effectiveness of our procedure, we survey many prior works and conduct experiments on the most two popular models (VGG-16 and ResNet-50) and dataset (ImageNet 2012). The experiment results show our proposed technique is able to reduce the computation requirement by 4.29X (VGG-16) and 2.84X (ResNet-50) while only sacrificing about 1.40\% top-5 and 2.50\% top-1 accuracies. Compared with many prior works, our results are much better in terms of both computation reduction and accuracies. However, when compared with some prior works, it may not be obvious to judge which one is better, for example, one pruned model requires less computation but has lower accuracies. In this case, our pruned models provide alternative data points in the design space and users can select the proper one to be deployed according to his/her specifications.

The remaining of this paper is organized as follows: we first give an overview of prior works in Section \ref{sec:priorworks}. In Section \ref{sec:notation_algo}, we introduce the notations that will be used throughout the paper and then present our channel pruning algorithm along with the tweaks for ResNet-like models. We show experiment results and the corresponding analysis in Section \ref{sec:results} followed by conclusion.
    \section{Overview of Prior Works}\label{sec:priorworks}

Model compression techniques can be categorized into four categories: \textbf{quantization}, \textbf{low-rank approximation}, \textbf{unstructured pruning}, and \textbf{structured pruning}. In this section, we briefly go over the techniques in the first three categories and then examine several structured pruning techniques in detail, which will be compared with our technique in later section.

\textbf{Quantization} aims to quantize and represent weights and/or activations with lower bit width. \cite{quantization} investigates the impact of quantization on both network accuracy and hardware metrics including memory footprint, power consumption, and design area. It demonstrates comprehensive tradeoff curves between accuracy and implementation cost under various quantization schemes. \cite{binary, ternary} push the boundary to the extreme where weights and/or activations are quantized into binary/ternary values, and thus achieve at least 16X model compression ratio. However, this impressive compression ratio cannot be easily translated into inference time speedup without the support of dedicated hardware or bitwise arithmetic libraries. Furthermore, specialized gradients computation and accumulation algorithm needs to be used during fine-tuning, otherwise the accuracy degradation may become unacceptable. 

\textbf{Low-rank approximation} aims to approximate a weight matrix with the product of lower-rank weight matrices, thus achieving reduction in both model parameters and computation. \cite{lowrank} applies Singular Value Decomposition (SVD) to higher order tensors by extending it to monochromatic approximation and biclustering approximation. But the compression ratios on convolutional layers are not as impressive as those on fully connected layers and the authors did not report reduction of computation and actual inference time. \cite{tucker} approximates weight matrices with Tucker decomposition and reports up to 4.93X computation reduction. However, low-rank approximation techniques effectively expand each layer to two or three layers, which adversely impacts the actual inference time depending on the platform. 

\textbf{Unstructured pruning} focuses on eliminating unimportant weights/connections in trained DNN models. These techniques can be dated back to 90s such as \cite{braindamage, brainsurgeon}, in which the second order derivatives of the loss function was used as a saliency measurement to determine if a weight should be pruned. In spite of its theoretical justification, high computational complexity is inevitable when applying to deep networks. Recently, the deep compression pipeline proposed in \cite{deepcompression} first prunes the weights with low magnitude, and then applies quantization and Huffman encoding to the pruned models. Although achieving up to 49X compression ratio, the time-consuming iterative pruning and fine-tuning is necessary to find the proper thresholds. \cite{dynamic} improves over \cite{deepcompression} by introducing mask variables and alternatively updating masks and model parameters. Thus, recovering the incorrect pruned weights and reducing training iterations. However, although unstructured weight pruning can greatly compress models' sizes with the help of sparse representation, the actual inference does not become faster without dedicated hardware or software libraries such as \cite{eie}.

On the contrary, \textbf{structured pruning} tries to prune models in a structured manner, such as pruning entire layers or groups of weights. Specifically, since the majority of inference energy is consumed by convolutional layers\cite{eyeriss} in CNNs, \textbf{pruning channels} of convolutional layers is a very effective technique to develop compact and efficient models. In earlier works, \cite{pruningFilters} prunes filters and the corresponding channels based on filters' $\ell_1$ norm. \cite{apoz} defines Average Percentage of Zeros (APoZ) to measure the percentage of zero activations of channels and neurons, and then prunes the channels/neurons with high APoZ values. \cite{taylorExpansion} first uses Taylor expansion to approximate the change of loss function with respect to pruning each channel, and then channels are pruned according to their impact to the loss function. \cite{ssl} proposes Structured Sparsity Learning (SSL), which imposes structured sparsity on model weights in terms of filters, channels, kernel shape, and depth during training. Thus, effectively pruning channels at the cost of lengthy training process.

More recently, \textbf{ThiNet}\cite{thinet} prunes the target layer by greedily selecting the input channel that has the least contribution to the output tensor in each iteration, and then this procedure is repeated until the specified number of channels have been pruned. The contribution of a input channel is defined as the squared summation of the corresponding partial output tensor values. The entire model is pruned layer by layer and fine-tuning is applied after pruning each layer. However, since all partial output tensor values are summed together to form the contribution value, this method may mistakenly treat a input channel as redundant even though its contributions to all output channels are relatively large but with near-to-zero summation.

For each layer to be pruned, \textbf{Channel Pruning}\cite{channelpruning} introduces a coefficient for each input channel, which will be multiplied with its coefficient to form the scaled input tensor. Then the problem of selecting redundant channels is formulated as a LASSO regression optimization problem with the goal of minimizing 1) the approximation error between the original and the new output tensor, which is obtained by using the scaled input tensor; and 2) the $\ell_1$ norm of the coefficient vector. The ratio between these two terms affects the number of zeros in the resulting coefficients, thus effectively determines how many channels will be pruned. Finally, the LASSO regression problem is solved by alternatively updating the coefficients and new weights until convergence. Similarly, \textbf{Network Slimming}\cite{networkslim} utilizes the scaling factors in Batch Normalization layers and trains the model under channel-level sparsity-induced regularization. After training, the channels with near-zero scaling factors will be pruned. However, besides the computationally expensive alternative updating/training process, it is also not clear how to directly set the ratio between the regularization term and the regular loss in these methods in order to prune any specified number of channels.

\textbf{Global Dynamic Pruning}\cite{gdp} shares the same alternative optimization scheme between the choice of channels to be pruned and model parameters. But instead of pruning models layer by layer, it globally and tentatively selects the channels to be pruned according to the Taylor expansion of the loss function with respect to each channel's weights. After the channels are selected, the parameters corresponding to the kept channels are fine-tuned. This two-step procedure is repeated until convergence, and finally, the selected channels are permanently pruned. However, since the redundant channels are globally selected across all layers, designers do not have fine-grained control over how many channels are pruned for each layer, which may lead to sub-optimal solution due to the lack of emphasis on pruning more beneficial (in terms of computation reduction) layers.
    \section{Notations and Our Algorithm}\label{sec:notation_algo}
In this section, we first introduce the basics and notations of channel pruning, then go through the details of our proposed technique.

\begin{figure}[t]
      \centering
      \includegraphics[width=3.4in]{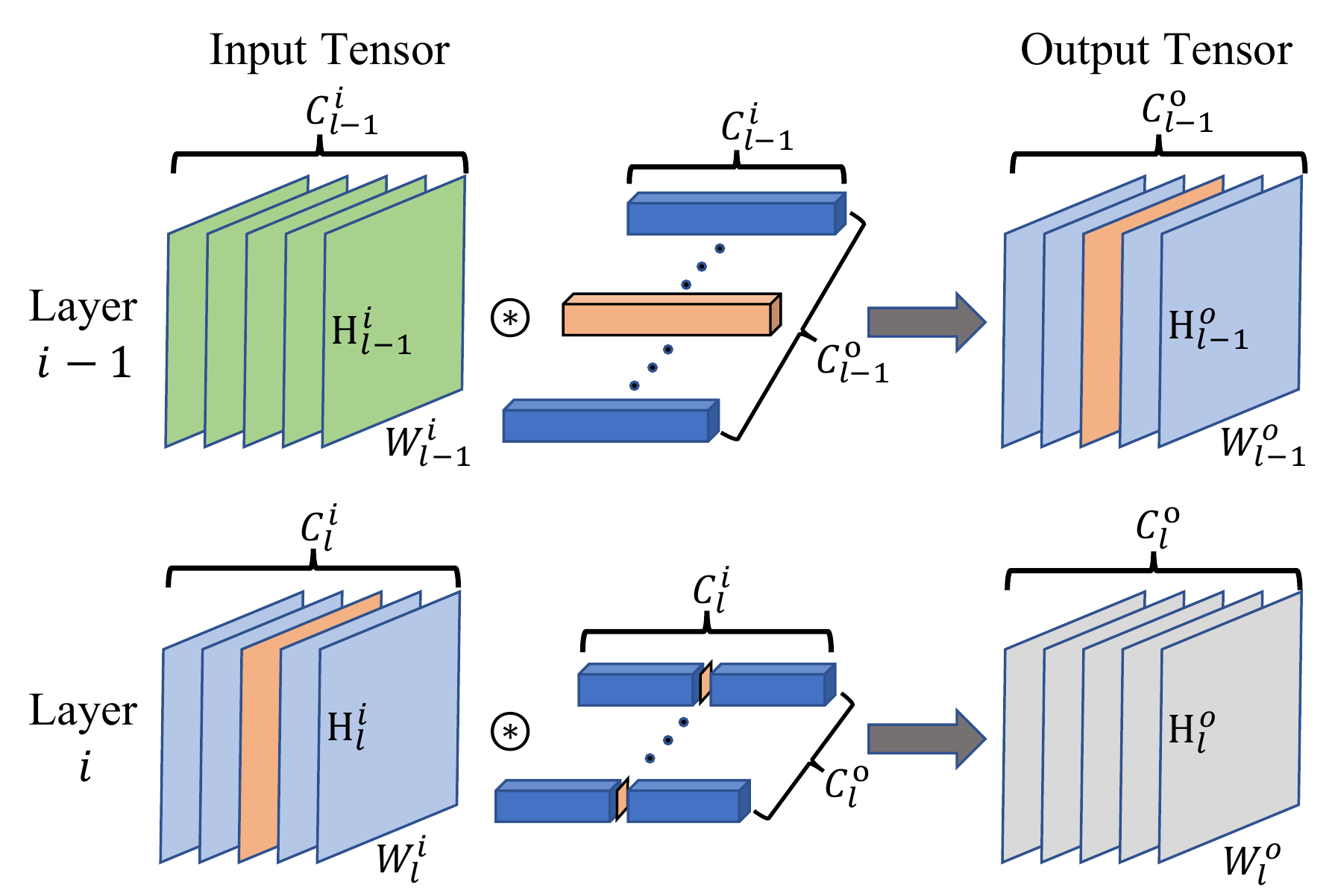}
      \caption{Diagram of convolutional layers in typical CNN models. The redundant channels and their corresponding weights are shown as orange slices and blocks, respectively.}
      \label{fig:conv_layer}
\end{figure}

\subsection{Basics and notations of channel pruning}

Fig. \ref{fig:conv_layer} shows two convolutional layers and their corresponding input and output tensors. For the sake of easy illustration, we assume batch size equals to 1 and thus ignore the dimension of batch size in the following discussion. The input to convolutional layer $l$ is a rank-3 tensor $\mathbf{I}_l \in \mathbf{R}^{H_l^i \times W_l^i \times C_l^i}$. The output rank-3 tensor $\mathbf{O}_l \in \mathbf{R}^{H_l^o \times W_l^o \times C_l^o}$ is obtained by performing convolution operations between the input tensor and layer $l$'s kernel $\mathbf{M}_l \in \mathbf{R}^{K_{l} \times K_{l} \times C_l^i \times C_l^o}$. Where, $H_l^{i/o}, W_l^{i/o}$, and $C_l^{i/o}$ are the height, width, and number of channels of the input/output tensors of layer $l$, respectively, and $K_l$ is the spatial dimension of layer $l$'s kernel. Then, $\mathbf{O}_l$ is subject to activation function and possibly max/average pooling to form the input tensor for the next layer. For each layer, the goal of channel pruning is to identify and prune redundant channels (among $C_l^i$ channels) of the input tensor $\mathbf{I}_l$ such that the overall accuracy does not degrade too much after pruning. Once the redundant channels are pruned, their corresponding weights in both $\mathbf{M}_l$ and $\mathbf{M}_{l-1}$ can be pruned as well. For example, if we can identify and prune $m$ channels in $\mathbf{I}_l$, then the resulting kernels of layer $l$ and $l-1$ will become $\mathbf{M}_l \in \mathbf{R}^{K_{l} \times K_{l} \times (C_l^i - m) \times C_l^o}$ and $\mathbf{M}_{l-1} \in \mathbf{R}^{K_{l-1} \times K_{l-1} \times C_{l-1}^i \times (C_{l-1}^o - m)}$\footnote{Usually $C_{l-1}^o = C_l^i$.}.

Although the ultimate goal is to maximally preserve the final output tensor during pruning, we propose to address this problem with a layer-by-layer approach. Thus, for each layer, the core of channel pruning is to identify and prune redundant channels of $\mathbf{I}_l$ and modify the remaining kernel weights, such that the output tensor $\mathbf{O}_l^{pruned}$ obtained by using only the remaining input channels $\mathbf{I}_l^{pruned}$ and modified weights $\mathbf{M}_l^{pruned}$ can maximally approximate the original output tensor $\mathbf{O}_l$. Mathematically, how well a tensor is approximated by another tensor can be described by the Frobenius norm of their difference. Thus, channel pruning can be formulated as:
\begin{align*}
    min  \ \ \ \ ||&\mathbf{O}_l - \mathbf{O}_l^{pruned}||_{\textbf{F}} \\
    where  \ \ \   &\mathbf{O}_l^{pruned} = \mathbf{I}_l^{pruned} \circledast \mathbf{M}_l^{pruned} \\
    s.t. \ \ \ \   &\mathbf{I}_l^{pruned} \subseteq \mathbf{I}_l \\
           \ \ \   &\mathbf{I}_l^{pruned} \in \mathbf{R}^{H_l^i \times W_l^i \times (C_l^i-m)}
\end{align*}
$m (<C_l^i)$ is the number of channels to be pruned, and $\circledast$ denotes convolution operation.

\subsection{Our proposed technique}\label{sec:algorithm}

\begin{figure}[t]
      \centering
      \includegraphics[width=3.4in]{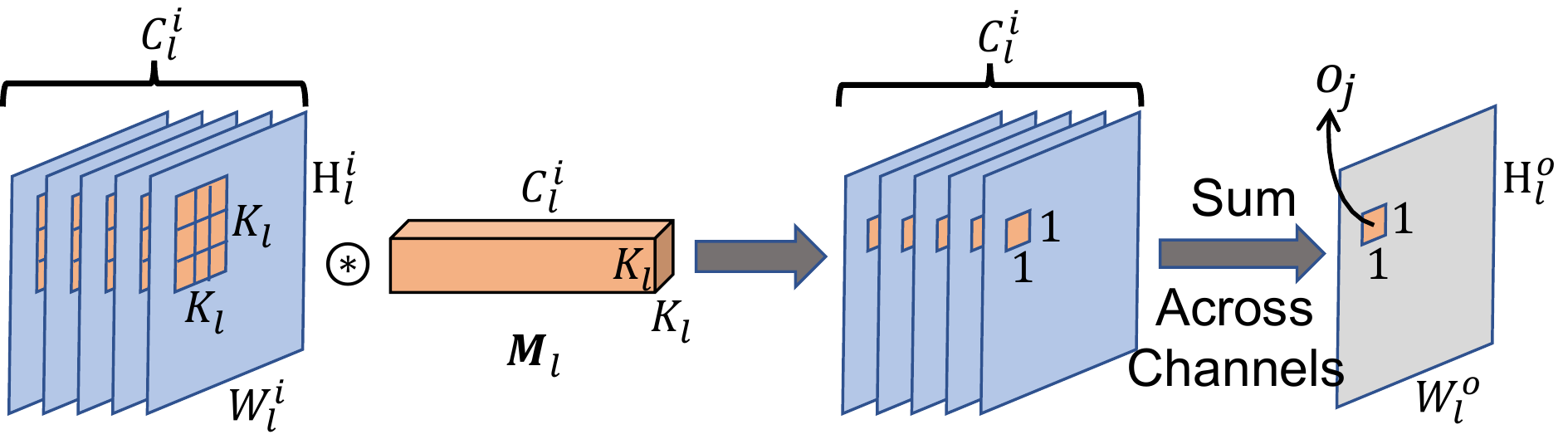}
      \caption{Diagram of obtaining a output channel by performing convolution operation between its kernel and the input tensor.}
      \label{fig:depthwise_conv}
\end{figure}

To solve the aforementioned problem, we investigate the linearity between the intermediate results of convolution operations and then identify redundant channels as those can be approximated by linear combinations of the other channels. By considering the intermediate results instead of the input tensor $\mathbf{I}_l$, we actually exploit the redundancy in both input tensor $\mathbf{I}_l$ and kernel $\mathbf{M}_l$ together. Specifically, for an input image, an element $o_j$ of the $j$th channel of $\mathbf{O}_l$ is obtained by:
\begin{equation}
    o_j = \sum_{c=1}^{C_l^i}\sum_{k_h=1}^{K_l}\sum_{k_w=1}^{K_l}\mathbf{I}_{l(k_h, k_w, c)} \times \mathbf{M}_{l(k_h, k_w, c, j)}
\end{equation}
Note that the bias term is ignored for clarity and does not effect the correctness of the algorithm. Denoting the contribution of $o_j$ by input channel $c$ as $o_j^c$, as shown in Fig. \ref{fig:depthwise_conv}, we have:
\begin{align}
    o_j &= sum(\overrightarrow{o_j^c})\\
    \overrightarrow{o_j^c} &= [o_j^1, o_j^2, ..., o_j^{C_l^i}]^T \\ 
    o_j^c &= \sum_{k_h=1}^{K_l}\sum_{k_w=1}^{K_l}\mathbf{I}_{l(k_h, k_w, c)} \times \mathbf{M}_{l(k_h, k_w, c, j)}
\end{align}

\begin{algorithm}[t]
\small
\caption{Find the most $C_l^i - m$ representative rows in $\mathbf{A}$}
    \label{alg:qr}
    \begin{algorithmic}[1]
        \Procedure{FindRepresentativeRows}{$\mathbf{A}$ , $C_l^i - m$}
            \State Perform SVD decomposition on $\mathbf{A}=\bU\bDelta\bV^T$
            \State Set $\bU_{C_l^i - m}$ to be the first $C_l^i - m$ columns of $\bU$ 
            \State Perform QRD on $\bU_{C_l^i - m}^T$ and get $\bU_{C_l^i - m}^T\bP=\bQ\bR$
            \State The first $C_l^i - m$ columns in permutation matrix $\bP$ identifies the most $C_l^i - m$ representative rows in $\mathbf{A}$
        \EndProcedure 
    \end{algorithmic}\vspace{-1mm}
\end{algorithm}

Thus, if we can identify a subset of $o_j^c$s of size $m$ from $\overrightarrow{o_j^c}$, such that each element in this subset can be approximated by linear combination of the $o_j^c$s outside of this subset, then these elements can be safely pruned and the original $o_j$ can be well-approximated by the summation of the scaled remaining $o_j^c$s. However, this is only for one element in one output channel with respect to one input image. In order to make this idea work for an entire convolutional layer, we need to consider $\overrightarrow{o_j^c}$ with three more dimensions:
\begin{enumerate}
    \item Consider $\overrightarrow{o_j^c}$ across all elements of the $j$th output channel, thus the redundancy is identified for the entire $j$th channel of the given input image.
    \item Consider $\overrightarrow{o_j^c}$ across all training samples, thus the redundancy is identified not only for one given image, but the entire training set.
    \item Consider $\overrightarrow{o_j^c}$ across all output channels in $\mathbf{O}_l$, thus the redundancy is identified for all output channels.
\end{enumerate}
With all of the three dimensions being considered, the vector $\overrightarrow{o_j^c}$ is extended to a matrix $\mathbf{A} \in \mathbf{R}^{C_l^i \times N}$, which is formed by concatenating $\overrightarrow{o_j^c}$ over $N$ samples. Ideally, $N$ is the number of $\overrightarrow{o_j^c}$s that can be collected from all $\mathbf{O}_l$s of all training images, but that is too big to be stored and processed. So, in practice, we perform random sampling to collect enough $\overrightarrow{o_j^c}$s from different images, output channels, and spatial locations. Since each row in $\mathbf{A}$ corresponds to an input channel, the problem becomes identifying $m$ redundant rows from $\mathbf{A}$ with the goal of best approximating the matrix $\mathbf{B} \in \mathbf{R}^{1 \times N}$ by linearly combining the remaining rows, where $\mathbf{B}$ is formed by extending $o_j$ over the samples. 

We propose to approximately and efficiently solve the inverse of this NP-hard subset selection problem, which is identifying the most $C_l^i-m$ representative rows in $\mathbf{A}$, by pivoted QR factorization as shown in Algorithm \ref{alg:qr}. Let's denote the remaining matrix after pruning the redundant rows as $\mathbf{A}_{pruned}$. To best approximate the matrix $\mathbf{B}$, the optimal scaling factors for the remaining rows, and thus their corresponding input channels, can be obtained by $\mathbf{B}\mathbf{A}_{pruned}^{\dagger}$, where $\mathbf{A}_{pruned}^{\dagger}$ is the pseudo-inverse of $\mathbf{A}_{pruned}$. Finally, the input channels and kernel weights correspond to those pruned rows can be eliminated, and the effect of scaling the remaining channels can be achieved by scaling the corresponding kernel weights, which effectively generates the new kernel weights.

\subsection{Handle ResNet architecture}\label{sec:handle_resnet}

\begin{figure}[t]
      \centering
      \includegraphics[width=3in]{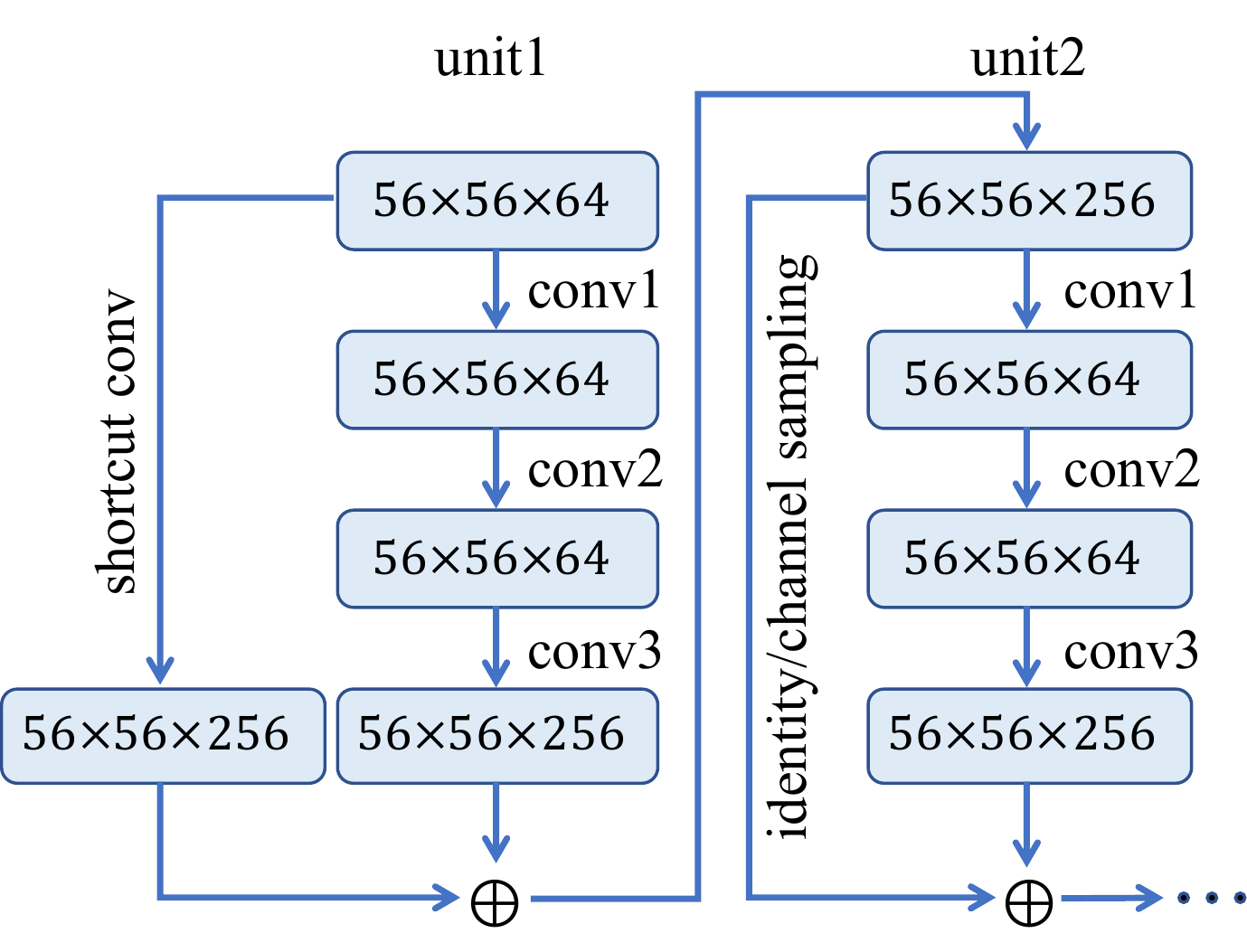}
      \caption{Diagram of the first two bottleneck units in ResNet-50 model for ImageNet classification task.}
      \label{fig:resnet}
\end{figure}

Fig. \ref{fig:resnet} shows the first two bottleneck units in ResNet-50 model, which is designed for ImageNet image classification task. Besides the typical pipelined convolutional layers in classic CNN models, it also has a shortcut path in each unit connects the input and output tensors of each unit. The shortcut path in \textit{unit1} is a projection path, which consists of a convolutional layer that projects $64$ input channels to $256$ output channels with the same spatial size. The shortcut path in \textit{unit2} is an identity path, which does nothing more than feed forward the input tensor to the output tensor. In prior works \cite{thinet, gdp, channelpruning} that have discussed the technical details about how ResNet's bottleneck unit is handled, only the input tensors of layer \textit{conv2} and \textit{conv3} are pruned, while the input tensor of layer \textit{conv1} is intact. This limitation is due to two reasons: 1) for bottleneck units with projection path, they were not able to analyze and prune layer \textit{shortcut conv} and \textit{conv1} simultaneously; and 2) for bottleneck units with identity path, pruning the input tensor of layer \textit{conv1} may cause misalignment problem when merging it with the output tensor of layer \textit{conv3}.

To overcome the above limitation and explore more pruning opportunities, we apply two tweaks to our technique to successfully prune all layers in ResNet architecture. First, for units have projection shortcut path, the redundant channels need to be identified for both layer \textit{shortcut conv} and layer \textit{conv1} at the same time. To handle this, when extending vector $\overrightarrow{o_j^c}$ to matrix $\mathbf{A}$ (refer to Sec. \ref{sec:algorithm} for details), we sample $o_j$ from the output tensors of both layer \textit{shortcut conv} and layer \textit{conv1}. Thus, matrix $\mathbf{A}$ has the information from both layers and the redundant input channels are identified for both layers simultaneously. 

Secondly, for units have identity shortcut path, the issue is that after pruning layer \textit{conv1}, the resulting input tensor may not be able to merge with the output tensor of layer \textit{conv3} due to misalignment between channels. Moreover, some channels that may \textbf{not} be identified as redundant by the next bottleneck unit can potentially be identified as redundant by the \textit{conv1} layer of the current unit and be pruned out, which is undesired. To handle this, we propose to prune layers in backward direction for the entire ResNet model, and propagate the set of indices of remaining input channels identified by the next unit back to the current unit. Later on, when pruning the \textit{conv1} layer of the current unit, another set of indices of remaining input channels will be identified, and the union of these two sets will finally be used as the remaining channels for the \textit{conv1} layer and propagated back to the previous unit. Since the final remaining input channels for layer \textit{conv1} is a superset of the channels that should be passed through the identity shortcut path, we perform \textit{channel sampling} to select the channels needed by the next unit and only pass them to the next unit.

    \section{Simulation Results}\label{sec:results}

\subsection{Experiment Setup}
After examining many prior works, we found that the most popular experimented models are VGG-16\cite{vgg} and ResNet-50\cite{resnet}, which are designed for and trained with the ImageNet ILSVRC 2012 dataset. So we experimented with these models and dataset in order to have a fair comparison with others. VGG-16 is a classic pipelined CNN model with 16 layers, which are divided into 5 convolution groups (each group has 2 or 3 convolutional layers) and 3 fully connected layers. ResNet-50 mainly consists of 4 bottleneck blocks with each of them has 3 to 6 bottleneck units. Each bottleneck unit has 3 convolutional layers and one identity/projection shortcut path depends on whether this unit is the first unit in its group. The original models used in this work are downloaded from TensorFlow's official website. ImageNet is a large-scale image dataset that contains 1.2M training images and 50k validation images of 1000 classes and various resolutions. Standard data augmentation schemes such as resizing, random cropping, and horizontal flipping are used during the fine-tuning of each model. All accuracies are obtained by using the single-view test approach (central crop only) on the validation set.

\subsection{Layers' Sensitivity Analysis}
\begin{figure}[t]
      \centering
      \includegraphics[width=3.4in]{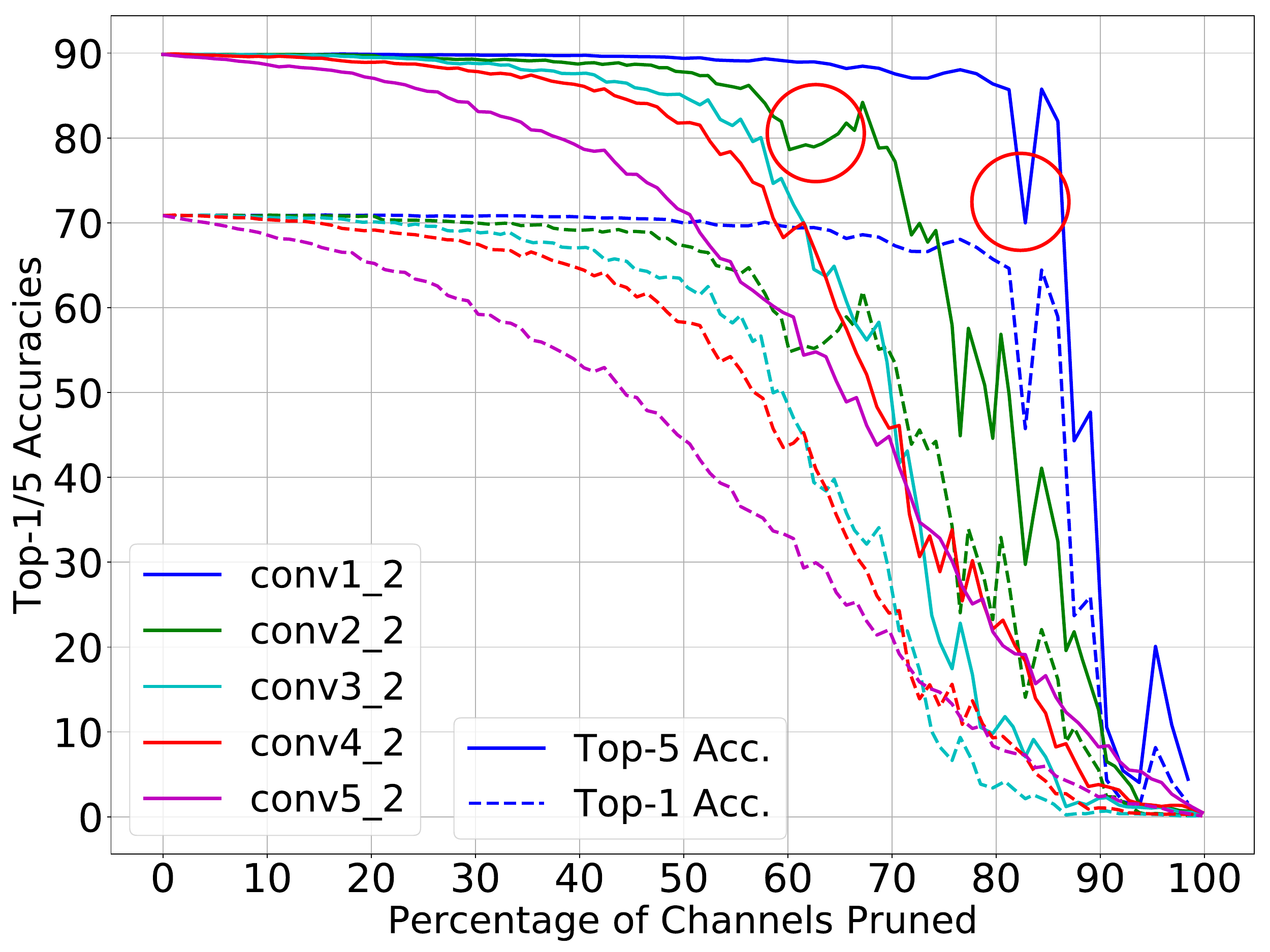}
      \caption{The sensitivities curves of the second convolutional layers of all 5 convolution groups in VGG-16 model under pruning.}
      \label{fig:vgg_trend}
\end{figure}

In order to verify the effectiveness of the proposed technique as well as determine to what extent each layer should be pruned when pruning the entire model, we performed sensitivity analysis by applying our technique to each layer independently for all layers in both VGG-16 and ResNet-50 models and tested how accuracies are impacted. 

Fig. \ref{fig:vgg_trend} shows how top-1/5 accuracies change when various percentages of input channels are pruned for several convolutional layers in VGG-16. Due to space limitation, we only show the sensitivity curves for 5 layers, which are the second convolutional layer of all 5 convolution groups. Since these layers have different number of input channels, the x-axis is normalized to show the percentage of channels being pruned instead of the absolute numbers. We make two observations from this figure: 1) Some layers, especially layer \textit{conv1\_2} and \textit{conv2\_2}, experience abnormal accuracies drops (circled in red) as their input channels are pruned. However, pruning less channels should always generate at least equally good result than pruning more channels, because we can simply set the weights of those extra channels to zero and obtain the same accuracies. Thus, these glitches are caused by the built-in randomness in the sampling procedure (refer to Sec. \ref{sec:algorithm} for details) of our algorithm, and they can be eliminated by either averaging the results from multiple runs or sampling more data points; 2) Comparing the curves between layers, it is obvious that early layers are more robust than later layers when the same percentage of input channels is pruned. This behavior is expected for typical CNN models like VGG-16, because early layers are responsible for extracting low-level features such as edges, colors, and corners of various orientations, and these features can easily be well-approximated by linear combinations of others. In contrast, features extracted by later layers are more complex and closely related to specific classes, thus it is harder to approximate the pruned channels with the remaining channels in these layers. 

\begin{figure}[t]
      \centering
      \includegraphics[width=3.4in]{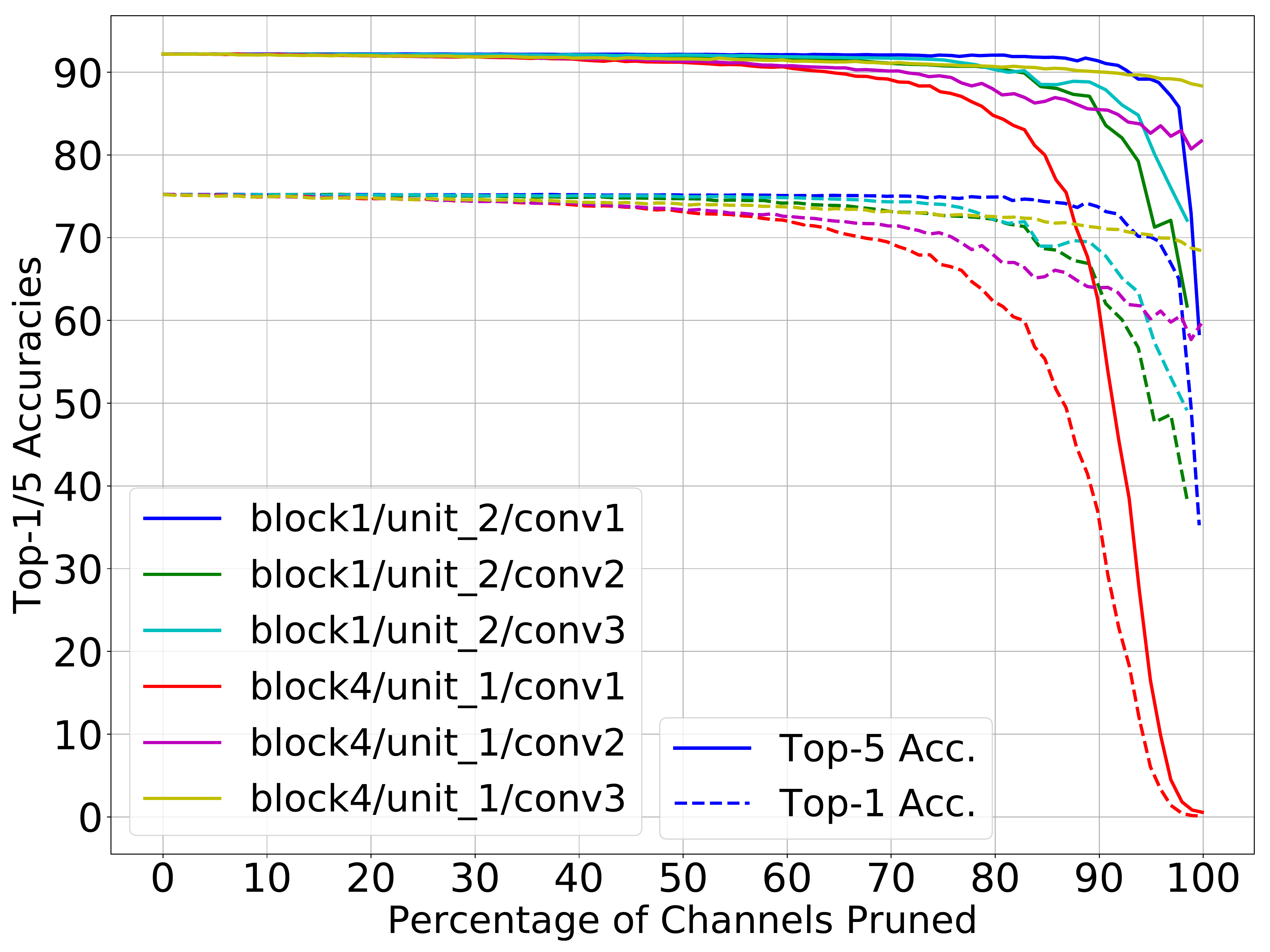}
      \caption{The sensitivities curves of the convolutional layers of two bottleneck units in ResNet-50 model under pruning.}
      \label{fig:resnet_trend}
\end{figure}

Fig. \ref{fig:resnet_trend} shows the sensitivity curves of 6 convolutional layers in ResNet-50 model. Due to space limitation, we choose to show 6 representative curves from two bottleneck units, which are \textit{unit\_2} from \textit{block1} and \textit{unit\_1} from \textit{block4}. From the figure, we make the following observations:

\textbf{First}, it is obvious that the sensitivity curves of the \textit{conv1} layers of \textit{block1/unit\_2} and \textit{block4/unit\_1} are dramatically different, the curve from \textit{block4} is much more sensitive than that from \textit{block1}. This difference is partially due to the reason we mentioned above, but more importantly, is because we handle the identity and projection shortcut paths differently. Since \textit{unit\_1} is the first bottleneck unit in \textit{block4}, it has a projection path, which is pruned simultaneously with layer \textit{conv1}. While \textit{unit\_2} is the second bottleneck unit in \textit{block1}, it has a identity path, which is intact to avoid the misalignment issue when pruning the \textit{conv1} layer. So it can still pass information to later layers. Therefore, when almost all input channels are pruned, accuracies drop to 0 for layer \textit{block4/unit\_1/conv1} but remain relatively high for layer \textit{block1/unit\_2/conv1}. 

\textbf{Secondly}, when comparing the curves of layers (except the \textit{conv1} layer of the first bottleneck unit in each block) within the same bottleneck unit, we observe that layer \textit{conv2} is almost always more sensitive than the others. This is because only \textit{conv2}'s kernel size is $3\times3$ and all the others are $1\times1$. Thus, when scaling the original kernel weights to get the new weights during pruning, the adjustments applied to \textit{conv2}'s weights are more coarse-grained than those to the other layers (refer to Sec. \ref{sec:algorithm} for details). This makes it harder to accurately approximate the pruned channels for \textit{conv2} layers, which in turn makes these layers more sensitive. This phenomenon is not observed from the curves of VGG-16 because all of its convolutional layers have $3\times3$ kernel. 

\textbf{Finally}, comparing the curves between VGG-16 and ResNet-50 models, we can see that the curves from ResNet-50 are much more robust than those from VGG-16 in general. This is because a ResNet architecture with $i$ bottleneck units actually has $2^i$ different paths from the input to output, which can be seen clearly by unrolling the network \cite{resEnsemble}. On the other hand, typical CNN models like VGG-16 only has one effective path. Thus, pruning one single layer (except the \textit{conv1} layer of the first unit in each block) of ResNet-50 only affects a subset of its paths, while the remaining unaffected paths can still contribute to high accuracies. However, pruning any layer in VGG-16 or any \textit{conv1} layer of the first unit in any block in ResNet-50 affects all paths, thus explains why accuracies can drop to 0 when pruning these layers.

\subsection{Results of Pruning the Entire Model}
\begin{table}[]
\caption{Comparison of top-1/5 accuracies, \# FLOPs, and average inference time of the original and various pruned VGG-16 and ResNet-50 models on ImageNet LSVRC 2012 dataset}
\label{tab:acc_comp}
\begin{tabular}{c|l|c|c|c|c}
\toprule
\multicolumn{2}{c|}{Model Name}
& \multicolumn{1}{c|}{\begin{tabular}[c]{@{}c@{}}Top-1\\ Accuracy\end{tabular}} 
& \multicolumn{1}{c|}{\begin{tabular}[c]{@{}c@{}}Top-5\\ Accuracy\end{tabular}} 
& \multicolumn{1}{c|}{\# FLOPs} 
& \multicolumn{1}{c}{\begin{tabular}[c]{@{}c@{}}Inference\\ Time\end{tabular}} \\
\midrule
\multirow{8}{*}{\rotatebox[origin=c]{90}{\textbf{VGG-16}}}
& Original                                              & 70.85\%    & 89.85\%    & 15.47B    & 208.25ms           \\
\cmidrule{2-6}
& ThiNet\cite{thinet}                                   & 69.80\%    & 89.53\%    & \ 4.79B   & -                  \\
& GDP\cite{gdp}                                         & 67.51\%    & 87.95\%    & \ 3.80B   & -                  \\
& CP\cite{channelpruning}                               & 67.80\%    & 88.10\%    & \ 3.52B   & -                  \\
& TE\cite{taylorExpansion}                              & -          & 84.50\%    & \ 4.00B   & -                  \\
& TE\cite{taylorExpansion} (\hspace{1sp}\cite{gdp})     & 65.20\%    & 84.86\%    & \ 4.20B   & -                  \\
& SSS\cite{sparseStructure}                             & 68.53\%    & 88.20\%    & \ 3.83B   & -                  \\
\cmidrule{2-6}
& Ours                                                  & 68.30\%    & 88.41\%    & \ 3.61B   & 111.43ms           \\
\midrule
\multirow{8}{*}{\rotatebox[origin=c]{90}{\textbf{ResNet-50}}}
& Original                                                      & 75.22\%    & 92.20\%    & \ 3.86B     & 105.90ms        \\
\cmidrule{2-6}
& ThiNet\cite{thinet}                                           & 71.01\%    & 90.02\%    & \ 1.71B     & -               \\
& GDP\cite{gdp}                                                 & 70.93\%    & 90.14\%    & \ 1.57B     & -               \\
& CP\cite{channelpruning}                                       & 72.30\%    & 90.80\%    & \ 2.60B     & -               \\
& SFP\cite{softfilter}                                          & 74.61\%    & 92.06\%    & \ 2.25B     & -               \\
& PF\cite{pruningFilters} (\hspace{1sp}\cite{sparseStructure})  & 72.88\%    & 91.05\%    & \ 1.54B     & -               \\
& PF\cite{pruningFilters} (\hspace{1sp}\cite{sparseStructure})  & 72.98\%    & 91.08\%    & \ 1.70B     & -               \\
\cmidrule{2-6}
& Ours                                                          & 72.74\%    & 90.88\%    & \ 1.36B     & \ 67.49ms       \\
\bottomrule
\end{tabular}
\end{table}

Based on the layers' sensitivity analysis, we set the prune target for each layer accordingly and prune the entire model. Specifically, for layers that are robust to pruning, we prune 50\% to 70\% of their input channels, and not pruning more even if accuracies are still high. This is because the sensitivity analysis is performed independently for each layer, and how previous layers are pruned affects the behaviors of later layers. So, instead of calculating the comprehensive interaction of sensitivity between layers, which is impractical due to the amount of computation required, we simply chose to left some margin for later layers. For layers that are sensitive to pruning, we prune up to 40\% of their input channels so that the accuracies not degrade too much. Like many prior works, after pruning each layer, we fine tune the pruned model to recover the lost accuracies. Table \ref{tab:acc_comp} shows the comparison of top-1/5 accuracies, number of FLOPs required per inference, and the average inference time of a minibatch of images for the original and various pruned VGG-16 and ResNet-50 models. 

For pruned VGG-16 models shown at the top part of Table \ref{tab:acc_comp}, Channel Pruning (CP)\cite{channelpruning} only reported speedup values instead of the absolute number of FLOPs required per inference, so we calculated it based on their released source code. Taylor Expansion based pruning (TE)\cite{taylorExpansion} only reported top-5 accuracy of the pruned model in their original paper, so we further included the numbers reported by \cite{gdp} for comprehensive comparison (the source of data is shown in parentheses). Compared to ThiNet's pruned model, ours requires 24.63\% less computation while sacrificing 1.50\% (top-1) and 1.12\% (top-5) accuracies. It is not very obvious to judge which pruned model is better, one might be chosen over the other depends on the target platform, accuracies, and runtime specifications. Compared to Sparse Structure Selection (SSS)\cite{sparseStructure}, our pruned model requires 5.74\% less computation while providing slightly higher top-5 accuracy and slightly lower top-1 accuracy. For all the other techniques, our pruned model always provides significantly higher top-1/5 accuracies with less computation requirement (except for CP, which is 2.5\% lower than ours). 

As for pruned ResNet-50 models shown at the bottom part of Table \ref{tab:acc_comp}, the statistics of the pruned models generated by CP and Pruning Filters (PF)\cite{pruningFilters} are calculated based on the released source code or borrowed from \cite{sparseStructure} due to similar reasons mentioned above. Compared to the pruned models from ThiNet, GDP, and CP, our pruned model requires up to 47.7\% less computation while still achieves significantly higher top-1/5 accuracies than all of them. This is partially due to the fact that our proposed technique for handling ResNet-like architecture can explore more pruning opportunities than those prior works. When compared with Soft Filter Pruning (SFP)\cite{softfilter} and PF, our pruned model still requires 11.7\% to 39.6\% less computation but at the cost of sacrificing 0.17\% to 1.18\% top-5 accuracy. These four pruned models (and potentially with many others) together form the tradeoff between computation requirement and model accuracies, and the choice of the proper model to be deployed depends on the target platform, accuracies, and runtime requirement.

The last column of Table \ref{tab:acc_comp} shows the averaged actual inference time of a minibatch of 64 input images. Since different GPUs and libraries were used in prior works, and also because TensorFlow version of those pruned models cannot be easily obtained, we only report the inference time of the original and our pruned models for a fair comparison. All inference times are measured with one Nvidia GTX 1080Ti GPU and averaged over the entire validation set. Compared with the theoretical 4.29X and 2.84X computation reduction, the actual speedup achieved on our GPU are only 1.87X and 1.57X, for pruned VGG-16 and ResNet-50 models, respectively. Since we observed that both CPU and PCIe interface are far from being fully utilized, we believe the gap between the theoretical and actual speedup values is mainly caused by the cache effect and memory accessing pattern in GPU, which is affected by the hardware itself, the network architecture, and TensorFlow library implementation. Thus, we anticipate that mobile and embedded systems may benefit more from the pruned models and such gap may be narrowed when performing inference on them due to their limited memory resources compared to powerful GPUs. 

    \section{Conclusions}

In this paper, we proposed 1) a pivoted QR factorization based channel pruning technique that is capable of pruning any specified number of input channels of any layer; and 2) two tweaks to explore more pruning opportunities in ResNet-like architectures. Using the proposed technique, we analyzed layers' sensitivity under pruning and demonstrated significantly better results than many prior works. In future, we plan to comprehensively explore the tradeoff between models' performance and cost by applying our technique to more CNNs, datasets, and platforms.

    \bibliographystyle{ieeetr}
    \bibliography{reference}
    
\end{document}